# Improving Automatic Summarization of Radiology Reports through Mid-Training of Large Language Models

Mengxian Lyu
Department of Health Outcomes and Biomedical Informatics
University of Florida
Gainesville, FL, USA
lvmengxian@ufl.edu

Cheng Peng
Department of Health Outcomes and Biomedical Informatics
University of Florida
Gainesville, FL, USA
c.peng@ufl.edu

Ziyi Chen
Department of Health Outcomes and Biomedical Informatics
University of Florida
Gainesville, FL, USA
chenziyi@ufl.edu

Mengyuan Zhang
Department of Health Outcomes and Biomedical Informatics
University of Florida
Gainesville, FL, USA
zhangm3@ufl.edu

Jieting Li Lu
Department of Engineering Education
University of Florida
Gainesville, FL, USA
jlilu@ufl.edu

Yonghui Wu
Department of Health Outcomes and Biomedical Informatics
Preston A. Wells, Jr. Center for Brain Tumor Therapy
University of Florida
Gainesville, FL, USA
yonghui.wu@ufl.edu

***Abstract*—Automatic summarization of radiology reports is an essential application to reduce the burden on physicians. Previous studies have widely used the "pre-training, fine-tuning" strategy to adapt large language models (LLMs) for summarization. This study proposed a subdomain adaptation through a mid-training method to improve clinical summarization. We explored three adaptation strategies: (1) general-domain pre-training, (2) clinical-domain pre-training, and (3) clinical-domain pre-training followed by subdomain (i.e., radiology) mid-training. We developed models using large-scale clinical text from the University of Florida (UF) Health and conducted mid-training and fine-tuning experiments using widely used benchmark datasets, including OpenI and MIMIC-CXR. The experimental results show that the mid-trained model, GatorTronT5-Radio, achieved the best performance, outperforming models without mid-training in both text-based measures (ROUGE-L) and factuality measures (RadGraph-F1). Our mid-training methods also demonstrate better few-shot learning and could alleviate the "cold start" problem reported in previous studies as a learning barrier. Our findings support the use of "pre-training, mid-training, fine-tuning," instead of the widely used direct fine-tuning strategy.**

## I. INTRODUCTION

Radiology reports serve as the primary way of communication in diagnostic imaging[1]. Radiologists spend a significant portion of their time synthesizing detailed "Findings" into concise "Impressions", which is a critical yet labor-intensive process, causing radiologist burnout[2]. Automatic summarization of radiology reports could help radiologists quickly locate important findings to alleviate the burden. While large language models (LLMs) have been widely adopted in automatic summarization, current approaches often rely on a "Pre-train then Fine-tune" strategy[3]. Models like GatorTron[4] or ClinicalT5[5] are typically pre-trained on broad biomedical corpora (e.g., PubMed abstracts) or heterogeneous clinical notes (e.g., discharge summaries, nursing notes). While this domain adaptation is effective for general medical tasks, it may not be optimal for radiology applications because of the variations between radiology and other medical sub-specialties.

This study aims to examine whether subdomain-adaption (i.e., radiology as a subdomain of medicine) could further improve the automatic summarization of radiology reports. Clinical documentation in different specialties often has different sections, documenting styles, and linguistic hierarchy[6]. For example, the language used in the admission notes or patient-provider dialogues is often characterized by complete sentences and a broad medical vocabulary. In contrast, radiology reports show a highly specialized style—marked by omitted function words, dense anatomical jargon, and frequent negation patterns—that differs from other types of notes in electronic health records (EHRs)[7]. We hypothesize that subdomain-adaptation of LLMs for radiology could better leverage radiology-specific language and structural priors for a better summarization performance. Such an LLM could outperform models trained on general-domain English text and general medical text for radiology applications.

In this study, we propose and evaluate a subdomain adaptation method through unsupervised large-scale pre-training, subdomain-specific mid-training, and supervised fine-tuning. Instead of directly fine-tuning a foundation LLM on the target task, we introduce an intermediate stage of training, i.e., mid-training, using subdomain-specific data. We compare our method with existing training methods for evaluation. We examine a general-purpose LLM, T5 model, which is widely used for automatic summarization, a clinical LLM, GatorTronT5, which is developed using over 90 billion tokens from UF Health and general English text, and our mid-training model, GatorTronT5-Radio, which is adopted from GatorTronT5 through mid-training using the MIMIC-CXR[8] dataset to capture radiology-specific syntax before downstream fine-tuning.

We evaluate models using the OpenI[9] (Indiana University Chest X-ray), a dataset from a different institution with out-of-distribution (OOD) samples. Therefore, we can better examine the generalizability of LLM-based summarization, where an LLM is pretrained at UF Health, mid-trained using MIMIC-CXR, and evaluated using the OpenI dataset. This study contributes a subdomain adaptation method for automatic clinical summarization of radiology reports through mid-training. The experimental results show that our subdomain adaptation method based on mid-training remarkably improved summarization, outperforming existing models in both lexical overlap (ROUGE[10]) and clinical factual consistency (RadGraph-F1[11]). Our subdomain-adapted models require significantly fewer samples to achieve comparable performance, indicating good few-shot learning performance.

## II. Related Work

### A. Automatic Radiology Report Summarization

Radiology report summarization is commonly formulated as *Findings → Impression* generation, where the goal is to condense detailed, often verbose findings into a short, clinically actionable impression. Early approaches to radiology report summarization relied largely on extractive methods that select salient sentences from findings[12]. Recently, large language models (LLMs) such as BART[13] and T5[14] have achieved state-of-the-art performance on benchmarks like MIMIC-CXR and OpenI. Despite these advances, existing work mainly focuses on architectural modifications or changing learning objectives applied directly during the fine-tuning stage[15]. These approaches typically fine-tune off-the-shelf LLMs directly using the target dataset without an intermediate stage of subdomain alignment[16].

### B. Clinical Domain Pre-training

Directly applying general domain LLMs to the medical domain presents significant challenges due to the "domain shift" phenomenon, where the distribution of medical vocabulary differs vastly from general web text[17]. A widely adopted solution is to adopt these general-purpose LLMs to the clinical/biomedical domain through continuous pre-training, where models are trained on large-scale in-domain corpora to acquire medical semantics before downstream fine-tuning. Representative examples include BioBERT[17] and GatorTron[4]. For text-to-text generation, SciFive[18] and ClinicalT5[5] extend the T5 architecture by continuing pre-training on biomedical literature and/or clinical notes. While domain-adapted models outperform general domain models, they are typically trained on heterogeneous data sources (e.g., web text, biomedical literature, clinical notes). As a result, such adaptation may not reflect specific linguistic patterns in radiology reports.

### C. Mid-training

Mid-training has recently been formalized as a distinct intermediate stage situated between pre-training and post-training, aimed at improving model capability and adjusting data distributions and training dynamics across phases[19]. Different from unsupervised pre-training, mid-training is typically framed as a targeted, transitional step: it often blends high-quality corpora (to maintain foundational linguistic robustness and stability) with specialized data formats (e.g., domain text, QA/instruction) introduced as a warm-up toward downstream objectives. In addition, mid-training is explicitly discussed as a warm-up stage for supervised post-training to enhance subdomain capability while keeping generalizability.

## III. Methods

### A. Task definition

We formulate radiology report summarization as a conditional sequence-to-sequence generation task. Let $X = (x_1, x_2, \ldots, x_n)$ denote the token sequence of the $Findings$ section, and $Y = (y_1, y_2, \ldots, y_m)$ denote the token sequence of the $Impression$ section. The objective is to learn a parameterized model $p_\theta(Y \mid X)$ that generates $Y$ conditioned on $X$, typically by maximizing the conditional log-likelihood $\sum_{t=1}^{m} \log p_\theta(y_t \mid y_{<t}, X)$.

### B. Data

**Mid-training dataset (MIMIC-CXR**[8]**)** is a large, publicly available, de-identified chest radiography dataset from Beth Israel Deaconess Medical Center (BIDMC) in Boston, covering imaging studies collected between 2011 and 2016. The database contains 377,110 images corresponding to 227,835 radiographic studies from 65,379 patients, and each study is accompanied by a contemporaneously written, semi-structured free-text radiology report. For the mid-training stage, we concatenate the "Findings" and "Impression" sections to form raw text sequences, enabling the model to learn the subdomain's lexical and syntactic patterns via unsupervised span corruption.

**Fine-tuning dataset (OpenI**[20]**)** (Indiana University Chest X-ray Collection) is a publicly accessible, de-identified dataset released via the U.S. National Library of Medicine's Open-i platform. The collection contains 3,955 radiology reports, where each report is typically associated with one or more images. The accompanying reports are semi-structured and commonly include multiple sections such as Indication, Comparison, Findings, and Impression, making the dataset suitable for report summarization and related generation tasks. We use the paired Findings and Impression sections for supervised fine-tuning.

Table I summarizes the hierarchical training strategy. Our pipeline consists of a from-scratch pre-training (>90B Tokens), a subdomain mid-training, and a supervised fine-tuning (3,955 reports).

TABLE I. COMPARISON OF DATASETS, DOMAIN SCOPE, AND TRAINING OBJECTIVES ACROSS THE HIERARCHICAL ADAPTATION PIPELINE.

| Training stage | Data Source | Domain | Data scale | Training objective |
|---|---|---|---|---|
| Clinical Pre-training | UF Health + PubMed, Wikipedia, MIMIC-III | Broad Clinical | >90B Tokens | Unsupervised |
| Mid-training | MIMIC-CXR | Radiology | 13M Tokens | Unsupervised |
| Fine-tuning | OpenI | Radiology (OOD) | 0.3M Tokens | Supervised |

## C. Model

**General-purpose LLM:** We use the **T5**[14] (Text-to-Text Transfer Transformer), a widely used model for summarization. T5 is an encoder–decoder Transformer pre-trained on the Colossal Clean Crawled Corpus (C4 Corpus). Pre-training is performed with a span-corruption denoising objective (masking contiguous spans and predicting the missing text), which encourages robust sequence modeling and has been shown to transfer effectively to summarization-style generation.

**Clinical LLM:** We pretrained **GatorTronT5**, a clinical foundation model based on the T5 encoder–decoder transformer architecture, using >90B-word corpus as our previous GatorTron[4], consisting primarily of de-identified UF Health clinical notes (>82B words) supplemented with PubMed, Wikipedia, and MIMIC-III text. We train three model scales (220M, 770M and 3B parameters) with a maximum context length of 512 tokens. Following the T5 framework, we adopt a span-corruption denoising objective, where contiguous text spans are replaced with sentinel tokens and the decoder is trained to reconstruct the masked spans. A domain-specific subworlds vocabulary is trained from scratch on the full corpus using a SentencePiece/BPE tokenizer.

**Subdomain-Adapted Model:** To bridge the gap between broad clinical knowledge and radiology specialization, we introduce **GatorTronT5-Radio**. This model is initialized from the pre-trained GatorTronT5 checkpoints and undergoes mid-training using the MIMIC-CXR dataset. By exposing the model to radiology-specific syntax and terminology via unsupervised span corruption before downstream fine-tuning, we aim to enhance domain alignment. We perform a mid-training using MIMIC-CXR to adapt the backbone model to radiology-specific language and structure before fine-tuning. Starting from the GatorTronT5, we continue training with the same T5-style denoising objective on radiology report text. To facilitate mid-training, we extract and concatenate the Findings and Impression sections to form raw radiology sequences, enabling the model to internalize radiology-specific lexical patterns, stylistic compression, and common negation/uncertainty constructions.

## D. Fine-tuning using OpenI

We fine-tuned and evaluated all models using OpenI for the summarization task using supervised *Findings → Impression* pairs. For each report, the Findings section is used as the input, and the Impression section is used as the target summary. In addition, OpenI serves as an external, out-of-distribution target dataset to MIMIC-CXR and the clinical pre-training corpus, allowing us to assess cross-institutional generalization. This fine-tuning stage aligns the mid-trained models to the specific summarization task.

## E. Few-shot learning

We conducted experiments to examine the few-shot learning of LLMs. We randomly sampled predefined numbers of samples from the OpenI training data. Specifically, we created subsets of size $k \in \{5, 10, 20, 30, 50, 100, 200, 500, 1000\}$. For each subset size $k$, we finetuned all models using identical hyperparameters. We examine the few-shot learning performance by comparing the performance by training using $k$ samples.

## F. Experiment Settings

We implemented all models in PyTorch using the Hugging Face Transformers library and conducted experiments on a high-performance computing cluster equipped with one NVIDIA B200 GPU.

For pre-training, we trained the model on the HiPerGator GPU cluster using distributed data parallelism with Megatron-LM model parallelism across 20 nodes × 8 GPUs/node (160 NVIDIA A100 80GB GPUs). Training used mixed precision (bf16) with the distributed optimizer enabled. Optimization used Adam/AdamW with a maximum learning rate of $1{\times}10^{-4}$ and weight decay 0.01. We applied a warm-up of 0.01 of total steps followed by annealing to a minimum learning rate of $1{\times}10^{-5}$. Training progress is monitored using held-out validation loss, and training is stopped upon convergence.

For mid-training, we initialized from the GatorTronT5 checkpoints and trained on MIMIC-CXR report text using the unsupervised span-corruption denoising objective. We constructed mid-training inputs by concatenating the Findings and Impression sections and truncating sequences to a maximum length of 512 tokens. Optimization used AdamW with a conservative maximum learning rate of $1 \times 10^{-6}$, with 200 warm-up steps followed by a linear decay schedule. The global batch size was set to 16, and models were trained for 2 epochs to expose the model to the radiology subdomain while mitigating catastrophic forgetting.

For downstream fine-tuning, we trained all model variants (General T5, Clinical GatorTronT5, and Subdomain-adapted GatorTronT5-Radio) using the OpenI dataset following the supervised sequence-to-sequence objective with Findings as input and Impression as the target summary to generate. We used a consistent learning rate of $5 \times 10^{-5}$ across all evaluated model sizes for controlled comparisons and set the effective batch size to 16. Inputs were truncated to 512 tokens, and target outputs were limited to 128 tokens. During evaluation, we generated summaries using beam search with beam=4 and seed=42.

### G. Evaluation

We evaluate models on both linguistic and clinical perspectives for comprehensive assessment. To quantify surface-level lexical similarity, we report ROUGE[10] and METEOR[21], which measure the n-gram overlap between generated impressions and reference summaries. Moving beyond exact string matching, we utilize BERTScore[22] to capture semantic similarity within the contextual embedding space, ensuring that meaning is preserved even when wording varies. We used RadGraph-F1[11], which measures factual consistency by extracting radiology entities and relations from generated and reference impressions and computing overlap-based F1.

## IV. Results

Table II compares performance on the OpenI dataset across three adaptation strategies—general-domain pre-training, clinical-domain pre-training, and clinical-domain pre-training + radiology mid-training—and three model scales (0.2B, 0.7B, 3B). Overall, we observe two consistent trends: (i) performance improves with larger model scale, and (ii) domain adaptation improves performance, with radiology mid-training providing the largest additional gains.

The scale-up law[23] holds for all strategies, where performance improves as model size scales up, with the 3B model achieving the best performance. We observed consistent performance improvement in pre-training, mid-training, and fine-tuning. Clinical pre-training provides a modest improvement over the general baseline at each scale, while mid-training contributes the largest performance improvements across ROUGE-L, METEOR, and BERTScore. For example, with 0.2B parameters, ROUGE-L increases from 0.4874 (T5-base) to 0.5018 (GatorTronT5-base) and further to 0.5281 (GatorTronT5-base-Radio). Similar findings were observed for the model with 0.7B parameters, where the radiology-adapted model achieves the best overall metrics. For GatorTronT5 with 3B parameters, GatorTronT5-XL-Radio achieved the best performance (ROUGE-L 0.6362 and BERTScore 0.8363).

The proposed mid-training strategy improved both surface similarity and clinical factual consistency measured by RadGraph-F1 for all models. For models with 0.2B and 0.7B parameters, RadGraph-F1 increases from the general baseline (0.4229, 0.4850) to the clinically pre-trained model (0.4416, 0.4969) and further improves after mid-training (0.4585, 0.5027). Our findings demonstrate the efficiency of subdomain alignment through mid-training for automatic summarization.

Figure I. Comparison of few-shot learning.

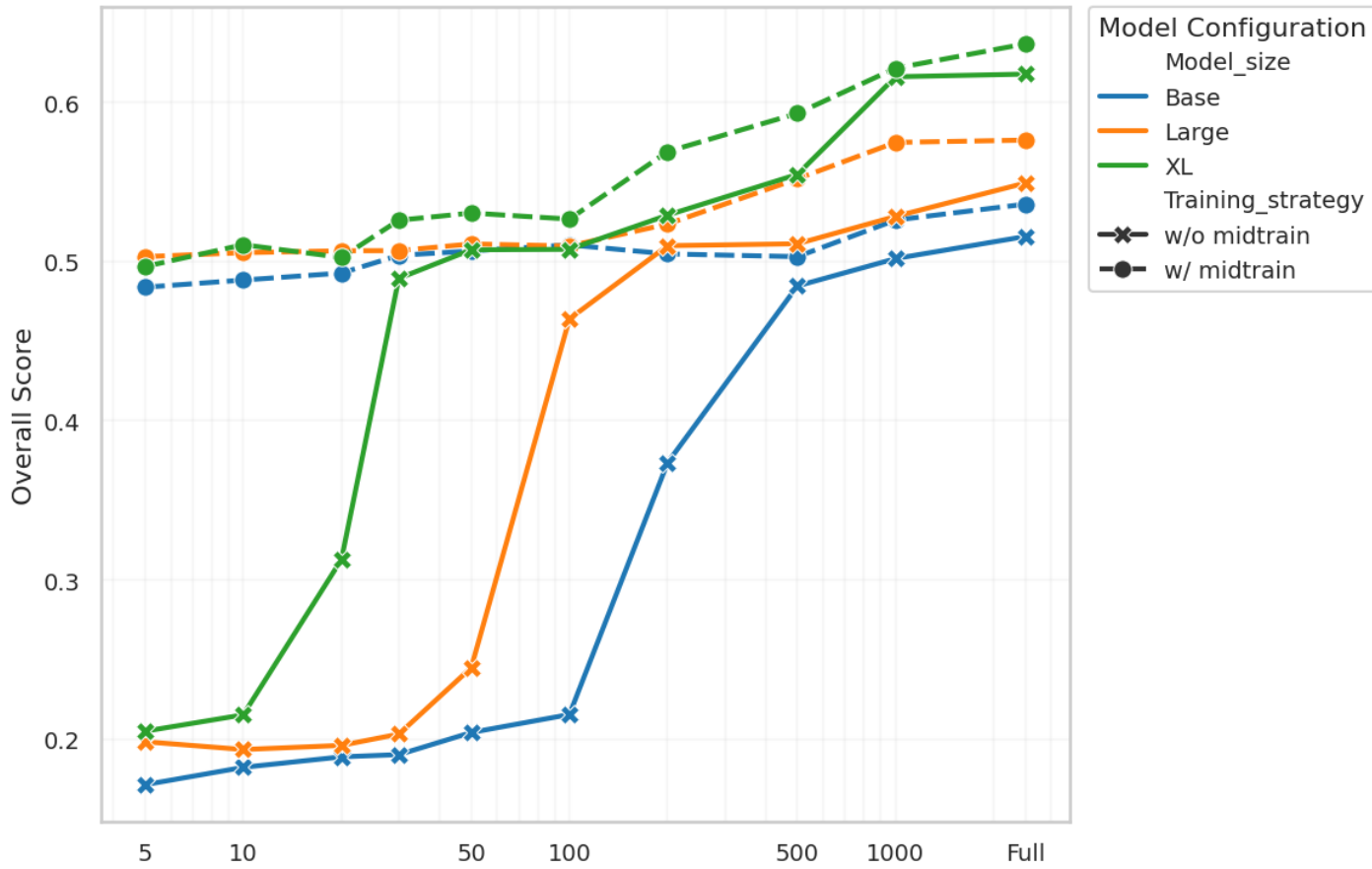

Figure I compares few-shot learning among all models. The comparison results show that mid-train models demonstrated better few-shot learning performance compared with their counterparts without mid-training, especially for scenarios with a very small number of samples (i.e., 5, 10). Models without mid-training show a "cold-start" learning pattern and struggle to learn from a small number of samples. For example, using only 5 training samples, GatorTronT5-Base without mid-training achieved ROUGE-L 0.0992, whereas the mid-trained counterpart reached 0.4716. Our findings suggest that mid-training could alleviate the "cold-start" learning obstacles in fine-tuning only models.

We also observed a threshold-like learning behavior[24] from models without mid-training: performance remains low until a critical amount of labeled data is reached, and this threshold decreases as model size increases. The Base model requires around 200 samples to show a substantial improvement (ROUGE-L from 0.17 to 0.34), while the Large and XL models reach comparable improvements earlier. This is consistent with our previous study[25] reporting that larger LLMs have better few-shot learning ability for information extraction. Our mid-training strategy remarkably alleviated the threshold effect across models of all sizes.

Table II Performance comparison of all models using the OpenI dataset

| Model | Model size | Adaptation strategy | Rouge-L | Meteor | BERTScore | Radgraph-F1 |
|---|---|---|---|---|---|---|
| T5-base | 0.2B | general pre-training | 0.4874 | 0.4547 | 0.7877 | 0.4229 |
| GatorTronT5-base | 0.2B | clinical pre-training | 0.5018 | 0.4645 | 0.7886 | 0.4416 |
| GatorTronT5-base-Radio | 0.2B | clinical pre-training + mid-training | **0.5281** | **0.4945** | **0.7991** | **0.4585** |
| T5-large | 0.7B | general pre-training | 0.5367 | 0.498 | 0.8021 | 0.485 |
| GatorTronT5-large | 0.7B | clinical pre-training | 0.5533 | 0.5124 | 0.8062 | 0.4969 |
| GatorTronT5-large-Radio | 0.7B | clinical pre-training + mid-training | **0.5709** | **0.534** | **0.8146** | **0.5027** |
| T5-XL | 3B | general | 0.6073 | 0.5712 | 0.8272 | 0.5495 |
| GatorTronT5-XL | 3B | clinical pre-training | 0.6157 | 0.5806 | 0.8294 | 0.5428 |
| GatorTronT5-XL-Radio | 3B | clinical pre-training + mid-training | **0.6362** | **0.6039** | **0.8363** | **0.5655** |

The well-known scale-up law holds for automatic summarization, where increasing model size improves the performance of summarization. The performance difference between models with and without mid-training narrows as more fine-tuning data becomes available, suggesting that

LLMs could learn subdomain knowledge with additional samples without mid-training. Nonetheless, mid-training models with all training samples outperform models without mid-training, indicating that mid-training captured additional knowledge that not exist in the training dataset. We observed similar findings in factual consistency as well: the best mid-trained model achieves RadGraph-F1 0.5655, compared with 0.4229 for the non-mid-trained Base baseline, supporting that subdomain adaptation improves both textual quality and clinical faithfulness.

## V. Discussion

Mid-training is a novel technique recently attracts many attentions to adopt LLMs for various domain-specific applications. Our findings demonstrate that mid-training is an efficient method to adapt a pretrained LLM for a specific application domain in radiology report summarization. Mid-trained models outperformed models without mid-training, indicating that mid-training is a necessary step before supervised fine-tuning. Our findings support the adoption of "pre-training, mid-training, and fine-tuning" adaptation, instead of directly fine-tuning pretrained LLMs.

Our subdomain adaptation based on mid-training could remarkably alleviate the "cold start" problem reported in previous studies. In few-shot learning scenarios (e.g., fewer than 200 samples), models without mid-training struggled to learn efficiently, whereas our mid-trained models show good few-shot learning ability. This "warm-start" capability is very useful for real-world deployment, where physicians prefer using minimal prompts to instruct LLMs for better performance.

Our findings suggest that a mid-trained small model can rival the performance of much larger models (e.g., XL) without mid-training, indicating the potential utility for low-resource settings. We observed performance improvements for both lexical similarity measures and clinical factuality measures.

## VI. Conclusion

This study presents a subdomain adaptation method based on mid-training. Our findings show that mid-training achieves better overall performance and few-shot learning performance and could remarkably alleviate the "cold start" learning problem. Our findings support the transition from direct fine-tuning to "pre-training, mid-training, and fine-tuning".

## Corresponding Authors

Yonghui Wu (yonghui.wu@ufl.edu) is the corresponding author.

## Acknowledgment

This study was partially supported by grants from the Patient-Centered Outcomes Research Institute® (PCORI®) Award ME-2023C3-35934, the PARADIGM program awarded by the Advanced Research Projects Agency for Health (ARPA-H), National Institute on Aging U24AG098157, National Institute of Allergy and Infectious Diseases, NIAID R01AI172875, National Heart, Lung, and Blood Institute, R01HL169277, R01HL176844, National Institute on Drug Abuse, NIDA R01DA050676, R01DA057886, R01DA063631, and the UF Clinical and Translational Science Institute. The content is solely the responsibility of the authors and does not necessarily represent the official views of the funding institutions.